\documentclass[article,accept,preprints,pdftex,moreauthors]{Definitions/mdpi}

\firstpage{1}
\pubvolume{1}
\issuenum{1}
\articlenumber{0}
\pubyear{2025}
\copyrightyear{2025}
\datereceived{ }
\daterevised{ } 
\dateaccepted{ }
\datepublished{ }
\hreflink{https://doi.org/} 

\usepackage{tabularx}
\usepackage{subcaption}
\usepackage[acronym]{glossaries}
\usepackage{xcolor}
\usepackage{framed}
\usepackage{rotating}

\glsdisablehyper

\definecolor{verylightgray}{gray}{0.97}

\newacronym{rnn}{RNN}{Recurrent Neural Network}
\newacronym{lstm}{LSTM}{Long Short-Term Memory network}
\newacronym{gru}{GRU}{Gated Recurrent Unit}
\newacronym{llm}{LLM}{Large Language Model}
\newacronym{bert}{BERT}{Bidirectional Encoder Representations from Transformers}
\newacronym{vqa}{VQA}{Visual Question Answering}
\newacronym{vsg}{VSG}{Visual Story Generation}
\newacronym{gpt}{GPT}{Generative Pre-trained Transformer}
\newacronym{bleu}{BLEU}{Bilingual Evaluation Understudy}
\newacronym{meteor}{METEOR}{Metric for Evaluation of Translation with Explicit ORdering}
\newacronym{cider}{CIDEr}{Consensus-based Image Description Evaluation}
\newacronym{nlp}{NLP}{Natural Language Processing}
\newacronym{dag}{DAG}{Directed Acyclic Graph}
\newacronym{ai}{AI}{Artificial Intelligence}
\newacronym{gpu}{GPU}{Graphics Processing Unit}
\newacronym{mt-bench}{MT-Bench}{Multi-turn Benchmark}
\newacronym{cnn}{CNN}{Convolutional Neural Network}
\newacronym{crcn}{CRCN}{Coherent Recurrent Convolutional Network}
\newacronym{fct}{FCT}{Fundação para a Ciência e Tecnologia}
\newacronym{brnn}{BRNN}{Bidirectional Recurrent Networks}
\newacronym{tapm}{TAPM}{Transitional Adaptation of Pretrained Model}
\newacronym{rouge}{ROUGE}{Recall-Oriented Understudy for Gisting Evaluation}
\newacronym{vwp}{VWP}{Visual Writing Prompts}
\newacronym{tf-idf}{TF-IDF}{Term Frequency-Inverse Document Frequency}
\newacronym{idf}{IDF}{Inverse Document Frequency}
\newacronym{fc}{FC}{Fully Connected}

\Title{Story Generation from Visual Inputs: Techniques, Related Tasks, and Challenges}
\TitleCitation{Story Generation from Visual Inputs: Techniques, Related Tasks, and Challenges}

\Author{Daniel A. P. Oliveira $^{1,2}$*\orcidA{}, Eugénio Ribeiro $^{1,3}$\orcidB{} and David Martins de Matos $^{1,2,}$\orcidC{}}\AuthorNames{Daniel A. P. Oliveira, Eugénio Ribeiro and David Martins de Matos}
\isAPAStyle{%
    \AuthorCitation{Oliveira, D. A. P., Ribeiro, E., \& Matos, D. M.}
}{%
    \isChicagoStyle{%
        \AuthorCitation{Oliveira, Daniel A. P., Eugénio Ribeiro, and David Martins de Matos.}
    }{
        \AuthorCitation{Oliveira, D. A. P.; Ribeiro, E.; Matos, D. M.}
    }
}
\address{%
    $^{1}$ \quad INESC-ID, Lisbon, Portugal; daniel.oliveira@inesc-id.pt (D.A.P.O.); eugenio.ribeiro@inesc-id.pt (E.R.); david.matos@inesc-id.pt (D.M.M.)\\
    $^{2}$ \quad Instituto Superior Técnico, Universidade de Lisboa, Lisbon, Portugal\\
    $^{3}$ \quad Instituto Universitário de Lisboa (ISCTE), Lisbon, Portugal}
\corres{Correspondence: daniel.oliveira@inesc-id.pt (D.O.)}

\abstract{Creating engaging narratives from visual data is crucial for automated digital media consumption, assistive technologies, and interactive entertainment. This survey covers methodologies used in the generation of these narratives, focusing on their principles, strengths, and limitations. The survey also covers tasks related to automatic story generation, such as image and video captioning, and visual question answering. These tasks share common challenges with \gls{vsg} and have served as inspiration for the techniques used in the field. We analyze the main datasets and evaluation metrics, providing a critical perspective on their limitations.}


\begin{document}
    \section{Introduction}\label{sec:introduction}
    In this document we survey the field of story generation from visual inputs, covering techniques, related tasks, and challenges.
    Stories are fundamental to human experience, serving as a bridge between imagination and reality~\cite{huhn2014handbook}.
    They represent the art of telling tales, whether real or imagined, and present a sequence of events that can captivate,
    inform, and provoke thought.
    Storytelling has been an essential part of human culture for millennia, serving as a means of communication, entertainment,
    education, and moral guidance.
    The art of storytelling has evolved over time,
    beginning with the practice of narrating stories through spoken words, advancing to written literature, and, more recently,
    transitioning to digital media formats.
    Stories have played a crucial role in shaping societies, passing down knowledge
    and values from one generation to another.
    Whether told around a campfire, written in books, or shared through
    videos and interactive platforms, storytelling continues to captivate and connect people, transcending time and
    technology~\cite{abbott_2008}.
    The field of story generation has gained attention,
    fueled by advances in \gls{nlp}~\cite{jurafsky2009} and computer vision~\cite{szeliski2010computer}.
    The goal is to create systems capable of generating novel and engaging stories,
    similar to those produced by humans~\cite{gervas_2009}.
    With the increasing availability of images and videos, there is a growing interest in generating stories based on this visual data~\cite{krishnaHRLN17}.
    We refer to this task as \gls{vsg}.

    \subsection{Definition and challenges}
    \label{subsec:definition-and-challenges-of-visual-story-generation}

    \gls{vsg} is the task of automatically creating coherent and engaging stories grounded on
    visual inputs, such as images or videos.
    It extends beyond simple image captioning or video description, requiring an understanding of complex relationships,
    temporal progression, and implicit context within the visual data.
    The generated narrative must, not only accurately represent the visual content, but also encapsulate the essence of the
    underlying story, thereby crafting an engaging experience for the audience.
    \gls{vsg} serves as a robust indicator of machine comprehension of visual content and poses several challenges,
    some of which are:
    \begin{itemize}
        \item \textbf{Understanding visual content:} The recognition of objects and scenes in the visual data,
        as well as understanding the relationships between them is crucial for generating coherent and meaningful narratives~\cite{xu2015ShowAA}.

        \item \textbf{Temporal coherence:} A story is inherently sequential and maintains a temporal order.
        Therefore, generated narratives should respect time progression, even if not in a linear manner (e.g., they may involve flashbacks, foreshadowing, or other such mechanisms), and ensure that events are narrated in a logically consistent manner.

        \item \textbf{Contextual understanding:} Often, elements of a story can be implicit or inferred from the context.
        Effective generation approaches must be capable of understanding and incorporating implicit information in the narrative.

        \item \textbf{Engagement and creativity:} One of the main goals of a story is to engage the audience.
        Therefore, the generated narrative should not only be accurate and coherent, but also creative and engaging.

        \item \textbf{Evaluation:} Determining the effectiveness of a \gls{vsg} system is a challenge in itself.
        Metrics originally developed for tasks such as machine translation and image captioning
        have been adapted for \gls{vsg}.
        However, these metrics are not appropriate to capture the creativity and engagement aspects of a story.
    \end{itemize}

    \subsection{Background and Motivation}\label{subsec:background-and-motivation}
    The motivations for \gls{vsg} are numerous.
    For example, \gls{vsg} can provide an interactive component to media consumption.
    Users could potentially influence the direction of the story by choosing different sequences of images or videos,
    creating a unique, personalized narrative.
    This could be particularly beneficial in areas like interactive gaming and movies, educational platforms, or even in digital
    marketing, where personalized narratives can greatly enhance user engagement and satisfaction.
    \gls{vsg} can also be used in journalism: it can be used to summarize large amounts of visual data, giving insights into the underlying patterns and relationships, and assisting in content creation.

    The growing interest in this area is also motivated by its potential to contribute to the understanding of human cognition and creativity. By developing models that emulate human-like storytelling abilities, researchers can gain insights into the cognitive
    processes involved in story creation, as well as explore the underlying mechanisms of creativity.
    These approaches can further lead to novel applications, such as more engaging virtual
    experiences~\cite{fan2018HierarchicalNS,peng-etal-2018-towards}.

    These use cases represent only a fraction of the potential applications of \gls{vsg}.
    As research in this field progresses, we anticipate that more applications will emerge, spanning an even broader range of domains.

    \subsection{Scope and structure of the survey}\label{subsec:scope-and-structure-of-the-survey}
    We start by defining the core elements of a story in Section~\ref{sec:the-elements-of-stories} as they are essential for understanding the structure and dynamics of a story.
    Section~\ref{sec:the-elements-of-stories} presents these elements and discusses their significance in storytelling.
    It also clarifies the distinctions between story, narrative, and plot, illustrating how a single story can be
    narrated in multiple ways to yield diverse experiences.

    In Section~\ref{sec:benchmark-datasets-and-evaluation-metrics} we then explore the main datasets used in story generation, and the various metrics used to assess their performance.
    We address the limitations that current evaluation metrics face in such an open-ended task and discuss potential
    directions for future research in this regard.

    The connection of \gls{vsg} with computer vision tasks such as \gls{vqa} and image/video captioning is also acknowledged,
    since they share core challenges, such as visual understanding and narrative generation.
    Thus, they have contributed with useful techniques to the field.
    Section~\ref{sec:from-nlp-to-computer-vision} explores this convergence between computer vision and \gls{nlp}.

    Concerning \gls{vsg} proper, we aim to present a balanced perspective on current methodologies, emphasizing opportunities
    and challenges for future research.
    The inclusion of a detailed analysis of deep learning methodologies with a broader review of additional and earlier techniques ensures broad coverage of the field.
    Section~\ref{sec:visual-story-generation} presents a broad perspective on the state-of-the-art of \gls{vsg}.

    Section~\ref{sec:real-world-applications} examines the real-world applications of \gls{vsg}.
    Finally, Section~\ref{sec:conclusions} presents the key findings of this survey and proposes directions for future research.

    \section{The Elements of Stories}\label{sec:the-elements-of-stories}
    Stories connect events and experiences through narration, ranging from personal tales to cultural epics.
    While the terms story, narrative, and plot are often used interchangeably, they have distinct meanings:
    a story is the chronological sequence of events (the ``what''), the narrative is how these events are
    presented (the ``how''), and the plot is the specific sequence as experienced by the audience~\cite{simpson2002oxford}.

    A single story can be told through multiple narratives, each offering different perspectives and experiences.
    Narratives may follow linear structures with chronological progression, or nonlinear approaches using flashbacks,
    flash-forwards, or parallel storylines that create complexity and require audience
    interpretation~\cite{virutal_reality_edu,mitchell2008cloud}.

    Story elements -- character, conflict, theme, setting, plot, and mode -- form the foundation for crafting
    and analyzing narratives.
    Understanding these elements is crucial for both human storytellers and automated story generation systems.
    In the following subsections, we detail each element and its significance in \gls{vsg}.

    \subsection{Character}\label{subsec:character}
    Characters are the entities through which stories unfold, with their decisions and actions propelling the plot forward~\cite{dibattista2011novel}.
    Through actions and speech, characters convey motives and emotions that shape audience perception and emotional connection.
    Initial character impressions influence audience expectations about future behavior.
    Characters who deviate from established patterns can create surprise or confusion, challenging audience understanding.
    Stories center around protagonists who encounter central conflicts, undergo personal growth, or acquire new knowledge.
    Antagonists add complexity and tension, often opposing the protagonist's goals.
    Traditional narratives feature heroes battling for moral causes against villains who perpetrate evil.
    Character development and evolution throughout the narrative engages audiences and drives story progression.
    The range of character types contributes to story depth and dynamics.

    \subsection{Conflict}\label{subsec:conflict}
    Conflict generates tension and drives character actions, shaping narrative trajectory and audience engagement~\cite{abbott_2008}.
    How conflicts are introduced, developed, and resolved determines story progression and emotional depth.
    Conflict resolution provides closure to the central problem and marks the culmination of narrative tension.
    Conflict encompasses any tension influencing characters, from the protagonist's main challenge to secondary internal struggles like anxiety or indecisiveness.
    These layers add complexity to character journeys and story depth.
    Longer narratives involve multiple conflicts that occur alongside the primary tension.
    Common conflict types include: Character vs. Character (opposition), Character vs. Nature (natural forces),
    Character vs. Society (social norms), Character vs. Unavoidable Circumstances (fate),
    and Character vs. Self (internal dilemmas).

    \subsection{Theme}\label{subsec:theme}
    Themes are the major underlying ideas of a story, often abstract and open to interpretation~\cite{griffith2010writing}.
    Unlike concrete story elements, themes invite discussion and allow for multiple interpretations and conclusions.
    They function as exploratory concepts that encourage audience reflection on deeper meanings.

    Themes can evolve as narratives unfold, revealing new dimensions and allowing audiences to develop personal interpretations of the story's messages.
    Different audience members may reach varying conclusions that differ from or exceed the creator's original intent.
    Understanding themes in narrative generation creates dynamic, thought-provoking storytelling experiences.

    \subsection{Setting}\label{subsec:setting}
    Setting encompasses the time, place, and context in which the story unfolds~\cite{truby2007anatomy}.
    It includes physical and temporal surroundings, as well as social or cultural conventions that influence character actions and decisions.
    Setting can act dynamically, possessing specific traits and undergoing changes that affect the plot.
    This allows it to evolve throughout the story, reflecting and impacting narrative progression.
    Hogwarts School in the ``Harry Potter'' series exemplifies an evolving setting that influences plot and character development~\cite{rowling2015harry}.

    \subsection{Plot}\label{subsec:plot}
    Plot is the sequence of events that propels the story from beginning to end~\cite{dibell1999elements}.
    Each plot point represents a moment of change that influences character understanding, decisions, and actions.
    This progression typically involves protagonists encountering conflict and attempting resolution, creating emotional stakes and suspense for both characters and audience.
    Plot advancement deepens audience investment in character fate.
    Stories can have multiple plots representing different event sequences or perspectives.
    Plot structure follows cause and effect, where character actions and events lead to reactions and consequences,
    forming a chain that moves the story forward.

    \subsection{Mode}\label{subsec:mode}
    Mode encompasses the choices and techniques authors use to frame narratives~\cite{pinault1992story}.
    A single story can be narrated using different modes, each offering distinct audience perspectives and experiences.
    Mode includes information scope, language style, medium, and the extent of narrative exposition and commentary.
    A key component is point of view or perspective.
    First-person narratives use personal pronouns like ``I'' and ``me,'' directly engaging audiences with the character's perspective.
    Third-person narratives avoid personal pronouns, offering more detached viewpoints.
    These perspectives exemplify narration techniques that explicitly tell stories through spoken or written commentary.
    While narration is common in written narratives, it can be optional in visual or interactive media.

    \section{Benchmark Datasets and Evaluation Metrics}\label{sec:benchmark-datasets-and-evaluation-metrics}
    This section presents and discusses benchmark datasets used for story generation and evaluation metrics that assess the performance of story generation models.
    Table~\ref{tab:story-datasets} provides a compilation of the datasets documented in the literature for \gls{vsg}.
    Among these datasets, VIST~\cite{huang-etal-2016-visual} is prominently utilized by the research community.
    This dataset serves as a primary benchmark, and most publications in this field report their experimental results on it.
    The VIST-Edit dataset~\cite{hsu-etal-2019-visual} provides human edited versions of machine generated stories,
    offering insights into human preferences for story quality.
    The Visual Writing Prompts dataset~\cite{tacl_a_00553} is noteworthy for its focus on stories that are visually grounded on the characters
    and objects present in the images.
    These datasets allow for the exploration of more complex and imaginative narrative structures, serving as a resource to train models to generate fictional stories.

    \begin{table}[htpb]
        \centering
        \begin{tabularx}{\textwidth}{p{4cm}|cX}
            \toprule
            Dataset                                                     & No. of Stories   & Notes                                                                                                                                 \\
            \midrule
            NY City~\cite{NIPS2015_17e62166}                       & 11,863          & 11,863 blog posts / 78,467 images                                                           \\
            \hline
            Disneyland~\cite{NIPS2015_17e62166}                    & 7,717           & 5 7,717 blog posts / 60,545 images                                                          \\
            \hline
            ROCStories~\cite{mostafazadehCHP16}                    & 100,000         & 5 sentences/story.                                                                          \\
            \hline
            LSMDC~\cite{rohrbachTRTPLCS16}                         & 118,114         & 202 movies.                                                                                 \\
            \hline
            VIST~\cite{huang-etal-2016-visual}                     & 50,000          & 210,819 unique photos                                                                       \\
            \hline
            VIST-Edit~\cite{hsu-etal-2019-visual}                  & 14,905          & Human-edited versions of 2,981 machine-generated stories                                    \\
            \hline
            Visual Writing Prompts~\cite{tacl_a_00553}             & 12,000          & 5 to 10 images/story, grounded on the characters.                                           \\
            \hline
            Pororo-SV~\cite{kimHCZ17}                              & 27,328          &                                                                                             \\
            \hline
            YouCook2~\cite{youcook-2}                              & 2,000           & Average length of 5.26 mins.                                                                \\
            \hline
            CMU movie summary corpus                               & 42,306          & Includes box office revenue, genre, release date, runtime, character names and actor names. \\
            \bottomrule
        \end{tabularx}
        \caption{Visual Story Generation Datasets.}
        \label{tab:story-datasets}
    \end{table}

    The evaluation metrics, used to assess the performance of \gls{vsg} models, provide
    quantitative measures of the quality of generated narratives.
    Most of these metrics were originally developed for machine translation or summarization but have been adopted by the community for other tasks such as
    image captioning and story generation: instead of comparing a generated text with a reference translation or summary, the generated text is compared with
    a reference caption in the case of image captioning, or with a reference story in the case of story generation.
    These metricsa include \gls{bleu}~\cite{papineni-etal-2002-bleu}, Self-BLEU~\cite{zhu2018TexygenAB},
    \gls{meteor}~\cite{banerjee-lavie-2005-meteor}, \gls{cider}~\cite{vedantamZP14a},
    \gls{rouge}~\cite{lin-2004-rouge}, Perplexity~\cite{jurafsky2023speech} and Bert Score~\cite{zhang2019BERTScoreET}.
    These metrics guide the development and refinement of \gls{vsg} models by offering insights into their linguistic fluency, textual diversity, and mostly lexical overlap with reference texts.
    They focus on specific linguistic and content-related aspects,
    providing quantitative measures for precision, recall, and consensus.
    However, they are limited by their focus on surface-level textual similarities rather than deep narrative alignment:
    since a story can be retold in various ways, employing different perspectives while preserving the original message,
    the corresponding generated text might deviate significantly from the reference text in terms of specific wording.

    Huang et al.~\cite{huang-etal-2016-visual} conducted an analysis
    to evaluate how well \gls{bleu} and \gls{meteor} correlate with human judgment in the context of \gls{vsg}
    according to different metrics.
    The results in Table~\ref{tab:human-assessment-correlation} show that
    \gls{meteor}, which places a strong emphasis on paraphrasing, has higher correlation with human judgment
    than \gls{bleu}.
    Nevertheless, in spite of \gls{meteor} showing the highest correlation with human judgment, this value is still low, suggesting that a large gap still exists between automatic metrics and human judgment.

    \begin{table}[htpb]
        \centering
        \begin{tabular}{l|cc}
            \hline
            Metric   & \gls{meteor}   & \gls{bleu}     \\
            \hline
            Pearson  & 0.22 (2.8e-28) & 0.08 (1.0e-06) \\
            Spearman & 0.20 (3.0e-31) & 0.08 (8.9e-06) \\
            Kendall  & 0.14 (1.0e-33) & 0.06 (8.7e-08) \\
            \hline
        \end{tabular}
        \caption{Correlations of automatic scores against human
        judgements, with p-values in parentheses~\cite{huang-etal-2016-visual}.}
        \label{tab:human-assessment-correlation}
    \end{table}

    To evaluate \gls{vsg} models, additional metrics that capture the creative aspects of storytelling may be necessary.
    These metrics might assess the uniqueness, novelty, and diversity of generated narratives,
    ensuring that models produce stories that go beyond mere data memorization.
    Coherence metrics could delve deeper into the logical flow of events, ensuring that
    generated narratives maintain a consistent and plausible storyline.

    The proliferation of diverse \glspl{llm},
    such as the GPT~\cite{openai2024chatgpt4o,openai2024devday,openai2024gpt4,openai2024gpt35}, Llama~\cite{touvron2023llama,llama3}, Mistral~\cite{mistral_mixtral,mistral_nemo_2024}, and Claude~\cite{anthropicSonnet4,anthropic2024claude21,anthropic2024claude2,anthropic2024introducingclaude}
    families, has led to the development of systematic scoring systems,
    aiming for objective assessments across a wide array of language understanding tasks.
    These scoring systems include \gls{mt-bench}~\cite{zheng2023judging} and Chatbot Arena~\cite{zheng2023judging} briefly described below.
    While they were initially developed for the evaluation of chatbots by human evaluators,
    they have been adapted to operate using a \gls{llm} as the evaluator achieving results
    close to those obtained by human evaluators.

    As \glspl{llm} continue to improve, they show potential in replacing human annotators in many tasks~\cite{gilardi2023ChatGPTOC}.
    Metrics to evaluate stories could be developed based on \glspl{llm}~\cite{zheng2023judging}, possibly following the underlying principles discussed above.

    \gls{mt-bench} is a benchmark tailored to test the multi-turn conversation and instruction-following capabilities of \gls{llm}-based chat assistants.
    It includes 80 high-quality multi-turn questions distributed across diverse categories such as writing,
    roleplay, reasoning, and knowledge domains.
    In the \gls{mt-bench} evaluation process, the user is presented with two distinct conversations, each generated by
    a different \gls{llm}-based chat assistant and is then tasked with deciding which assistant, A, B, or
    indicating a tie, better followed the instructions and answered the questions.
    For \gls{vsg} evaluation, this framework could be adapted by using an \gls{llm} as an automated judge to compare
    stories generated by different models, evaluating them based on criteria such as narrative coherence, visual grounding, and creativity.

    Chatbot Arena introduces a crowd-sourcing approach, fostering anonymous battles between \gls{llm}-based chatbot models.
    Users interact with two anonymous models simultaneously, posing identical questions to both and subsequently voting for the
    model that provides the preferred response.
    The Elo rating system~\cite{zheng2023judging} is employed to rank the models based on their performance in these battles.
    This rating system assigns a numerical score to each model based on their battle outcomes,
    adjusting these scores higher or lower after each encounter to reflect their relative ability to satisfy user queries effectively compared to their opponents.
    A similar automated evaluation system for \gls{vsg} could employ an \gls{llm} judge to assess pairs of generated stories
    given the same image sequence, ranking models based on story quality across multiple dimensions without requiring human annotators.

    In summary, while current metrics provide insights into specific dimensions of \gls{vsg},
    the development of new metrics may be necessary to holistically evaluate creativity, coherence, and emotional engagement,
    ensuring a more comprehensive understanding of the capabilities and limitations of these models.
    We anticipate that \glspl{llm} can play a role in the development of such metrics, serving as automated judges that
    evaluate stories based on multiple criteria.

    \section{From NLP to computer vision}\label{sec:from-nlp-to-computer-vision}
    \glsreset{vqa}\glsreset{nlp}
    This section provides an overview of areas at the intersection of computer vision and \gls{nlp} that are relevant to the topic of \gls{vsg},
    namely image and video captioning and \gls{vqa}.
    These tasks share a common challenge: they require the understanding of visual content and the ability to generate text based on that visual content.
    Some of the techniques used in these tasks have been adapted to \gls{vsg}.

    \subsection{Image and Video Captioning}
    \label{subsec:image-and-video-captioning}

    The transition from image and video captioning to \gls{vsg} represents a shift from describing individual images or video segments to narrating image sequences or video segments.
    While the former focus on \textsl{what} is in images or videos, the latter also considers \textsl{why} and \textsl{how}, capturing the underlying narrative that links a series of images or video frames.

    \gls{vsg} models need to understand the temporal progression of events,
    causal relationships, and the ability to construct cohesive and engaging stories.
    \gls{vsg} can also speculate on the motivations and emotions of characters
    since such interpretations do not contradict the visual content.
    Therefore, they build upon the foundational principles of image and video captioning, incorporating both the
    interpretation of visual data and the generation of coherent, contextually appropriate narratives.

    The tasks of image and video captioning consist in generating descriptive, human-readable sentences that represent the content of an image or video,
    identifying the main components in a visual scene and their relationships,
    alongside the ability to express these details in natural language.

    One notable contribution in this domain was the application of encoder-decoder architectures, known as the `show and tell'
    approach~\cite{vinyalsTBE14, yao2015describing, anderson2017BottomUpAT}.
    This technique employs \glspl{cnn} as encoders to extract visual features.
    Specifically, 2D-\glspl{cnn} are employed for image processing where the model interprets two dimensions: height and width.
    On the other hand, for video processing, 3D-\glspl{cnn} are typically used where an additional dimension,
    time or depth, is taken into account to perceive the temporal dynamics in video sequences.
    Then, \glspl{rnn} or their improved variants, \glspl{lstm}~\cite{hochSchm97}
    and \glspl{gru}~\cite{chungGCB14}, known for their ability to effectively model sequential data, are commonly
    used as decoders in image and video captioning tasks, enabling the generation of coherent and
    contextually meaningful textual captions and descriptions from the visual features extracted by the \glspl{cnn}.
    This combination of \glspl{cnn} and \glspl{rnn}
    effectively bridges the gap between vision and language.

    In the realm of image captioning, Transformers~\cite{vaswani2017AttentionIA} have been used as encoders to
    process visual features extracted from images.
    Leveraging the self-attention mechanism inherent in Transformers, these models are able to capture
    relationships between visual elements, attending to relevant regions in the image, and highlighting visual
    cues and their interdependencies~\cite{olimov2021ImageCU, radford2021LearningTV, liu2021SwinTH, yu2019MultimodalTW}.
    Transformer models also demonstrate the ability to generate
    precise and elaborate textual descriptions, significantly improving the overall quality of image captions.

    Similarly, Transformers have made contributions to video captioning, addressing the challenges associated
    with understanding and describing temporal dynamics in videos.
    By treating video frames as a sequential series of images, Transformers can leverage their attention mechanism to capture
    long-range dependencies across frames~\cite{liu2021VideoST, lin2021SwinBERTET}
    This enables the models to map the temporal progression of events, infer causal relationships, and grasp the
    underlying narrative structure within the video.

    The integration of Transformers in image and video captioning has yielded improvements that surpass the performance of traditional encoder-decoder architectures~\cite{liu2021SwinTH, liu2021VideoST}.
    Transformers not only generate semantically accurate captions but also exhibit enhanced coherence and contextual relevance.
    Their attention mechanism allows for a incorporation of local and global context, enabling a more
    holistic understanding of the visual content and facilitating caption generation.

    The use of Transformers in image and video captioning holds great promise, but there are challenges to be addressed.
    Efficient fusion of visual features with textual information~\cite{radford2021LearningTV}, handling long-term dependencies in video sequences~\cite{liu2021VideoST}, and ensuring temporal coherence in caption generation for longer videos are among the ongoing research topics.

    \subsection{Visual Question Answering}
    \label{subsec:visual-question-answering}
    \glsreset{vqa}

    \gls{vqa} combines computer vision and \gls{nlp} to enable the extraction of information from visual content, identifying objects, scenes, actions, and comprehending context, and provide meaningful answers to questions about that content.
    Both \gls{vqa} and \gls{vsg} involve understanding visual content and generating meaningful responses and the developments in \gls{vqa} can be directly applied in \gls{vsg} tasks.
    Through the use of \gls{vqa} techniques, \gls{vsg} models can gain valuable insights into the content and context of images and videos, improving the overall quality of their visual narratives.

    Encoder-classifier deep-learning \gls{vqa} models encode
    visual and textual inputs and select an answer from a predefined set using a classification process. Image encoders, typically pretrained \glspl{cnn}, extract visual features from the input images. Question encoders, often \glspl{rnn}, processes the input questions, converting them into fixed-length latent representations.
    Fusion techniques combine the visual and textual features, creating a fused representation that captures their interactions that is then classified using fully connected layers, mapping them to possible answer classes. Traditional \gls{vqa} models are trained using supervised learning with large-scale datasets, where image-question-answer triplets are provided as training examples.
    The parameters are optimized to minimize the loss between predicted answers and ground truth answers during training. Some examples are BUTD~\cite{anderson2017BottomUpAT}, that uses an attention-based model that combines bottom-up and top-down mechanisms to improve performance in image captioning and visual question answering tasks, and ``Show, Ask, Attend, and Answer''~\cite{kazemiE17}, that combines \glspl{lstm} with multiple attention layers.

    The transition from encoder-classifier methods
    to Transformer methods in \gls{vqa} was driven by the Transformer's ability to capture long-range dependencies and global contextual information.
    Transformers offered the potential to more efficiently capture the interactions between visual and textual modalities, enabling a more comprehensive understanding of the relationship between images and questions.
    In these \gls{vqa} models, images and questions are encoded using Transformer encoders and self-attention mechanisms capture the dependencies between words and the visual features, allowing the model to focus on relevant image regions while processing the question. The fused encoded representations are then fed into
    classification layers to generate answers~\cite{neurips_d46662aa,wang2022image,wang2022OFAUA,wang2023onepeace}.
    Recent advances in \gls{vqa} also explore new models that approach the task as a generation problem rather than a classification one, allowing for a more flexible and diverse range of answers~\cite{chen2023pali}.

    \section{Visual Story Generation}\label{sec:visual-story-generation}
    \gls{vsg} combines computer vision and \gls{nlp} to create coherent narratives from sequences of images or videos.
    This task extends beyond simple image captioning by requiring models to understand temporal relationships, infer causal connections between visual scenes, and generate engaging stories that capture the underlying narrative flow.
    Unlike video captioning, which focuses on objective description of visual content, \gls{vsg} can also speculate on the motivations and emotions of characters since such interpretations do not contradict the visual content.

    The field has evolved from early encoder-decoder architectures using \gls{cnn} and \gls{rnn} to Transformer-based models that leverage large-scale pretraining.
    These approaches must address challenges including visual understanding, temporal coherence, and narrative creativity while maintaining grounding in the provided visual content.
    The following subsections present the key methodologies and their contributions to advancing \gls{vsg} capabilities.

    \subsection{Visual Storytelling with Convolutional and Recurrent Neural Networks}
    \label{subsec:visual-storytelling-with-convolutional-and-recurrent-neural-networks}
    Huang et al.~\cite{huang-etal-2016-visual} proposed a dataset of sequential vision-to-language mappings, SIND v.1 (later renamed to VIST), with 81,743 images in 20,211 sequences, with both descriptive (captions) and story language, and aiming at facilitating research, from understanding individual images to comprehending image sequences that narrate events over time.

    This work also proposed models for generating stories based on image sequences, using a sequence-to-sequence \gls{rnn} approach, which extended the image captioning technique. It extracts features from each image using a pretrain \gls{cnn} called AlexNET~\cite{NIPS2012_c399862d}. It then encodes image sequences by using a \gls{gru} over the features of each individual image. The story decoder model consists of another \gls{gru} and generates narratives word by word.
    Four different decoding methods, shown in Table~\ref{tab:visual-storytelling-baselines}, were considered: beam search, greedy search, greedy search with removal of duplicates, and greedy search with removal of duplicates and incorporation of grounded words.

    \begin{table}[htpb]
        \centering
        \begin{tabular}{l|c}
            \hline
            Method                         & \gls{meteor} Score        \\
            \hline
            Beam=10                        & \cellcolor{red!20}23.13   \\
            Greedy                         & 27.76                     \\
            Greedy - Duplicates            & 30.11                     \\
            Greedy - Duplicates + Grounded & \cellcolor{green!20}31.42 \\
            \hline
        \end{tabular}
        \caption{
            Results of each decoding approach using evaluated using \gls{meteor} scores on the VIST~\cite{huang-etal-2016-visual} dataset.
            Red indicates the lowest score, green the highest.
        }
        \label{tab:visual-storytelling-baselines}
    \end{table}

    Results indicate that beam search alone does not perform well in generating high-quality stories.
    This contrasts with the results for image captioning, where beam search is more effective than greedy search.
    The reason is that generating stories with beam search results in generic high-level descriptions such as ``This is a picture of a dog'' originating from the label bias problem.
    On the other hand, by using a beam size of 1 (greedy search) there is an improvement in the quality of the generated stories. 

\begin{figure}[htpb]
        \centering
        \includegraphics[width=\textwidth]{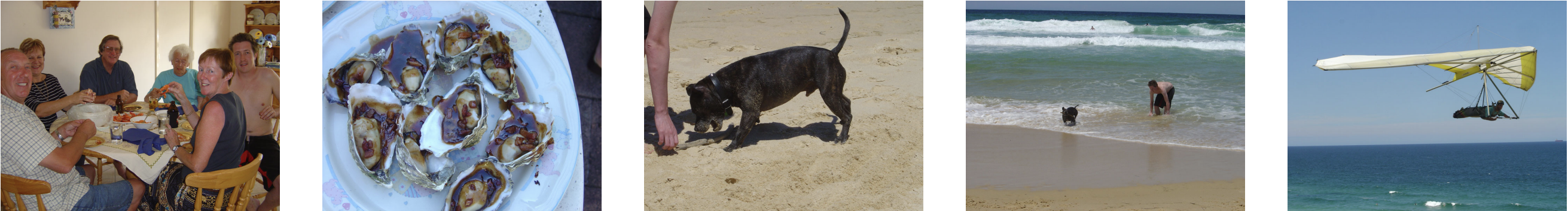}
        \caption{Sequence of images used to generate stories using different decoding methods.
        \textbf{Beam search:} This is a picture of a family. This is a picture of a cake.
        This is a picture of a dog. This is a picture of a beach. This is a picture of a beach.
        \textbf{Greedy:} The family gathered together for a meal. The food was delicious. The dog was excited to be there.
        The dog was enjoying the water. The dog was happy to be in the water.
        \textbf{Greedy - duplicates:} The family gathered together for a meal. The food was delicious. The dog was excited to be there.
        The kids were playing in the water. The boat was a little too much to drink.
        \textbf{Greedy - duplicates + grounded words:} The family got together for a cookout. They had a lot of delicious food.
        The dog was happy to be there. They had a great time on the beach. They even had a swim in the water.
        Adapted from~\cite{huang-etal-2016-visual}.
        }
        \label{fig:visual-storytelling-example}
    \end{figure}

    Since stories generated using a greedy search may sometimes contain repeated words or phrases that can adversely affect textual quality, a duplicate-removal heuristic was added to ensure that content words cannot be produced more than once on a given story. 
    Finally, to further improve quality, the previous approach was extended to include ``visually grounded'' words.
    These are words that are authorized for use by the story model only if they are licensed by the caption model:
    for a word to be included in the story, it must have appeared with some frequency in the captions generated
    by the caption model for the sequence of images.
    Fig.~\ref{fig:visual-storytelling-example} shows an example of the generated stories for an image sequence.

    \subsection{Visual Storytelling with Convolutional and Bidirectional Recurrent Networks Networks}
    \label{subsec:visual-storytelling-with-convolutional-and-recurrent-networks}

    Park et al.~\cite{NIPS2015_17e62166} proposed
    \gls{crcn}, integrating various neural network components
    to bridge the gap between visual content and textual narrative:
    VGGNet~\cite{simonyanZ14a} as image encoder and \gls{brnn} to capture contextual information from both the forward and backward directions within a given text sequence.
    \gls{crcn} also introduces a local coherence model that focuses on maintaining coherence and context within the generated sentence sequences.
    It also utilizes parse trees to detect grammatical roles and structural attributes,
    ensuring that the generated sentences integrate with the broader narrative context.
    During training, correctly aligned image-sentence pairs are given higher scores than misaligned pairs.

    The evaluation of \gls{crcn} considered several automatic metrics and several baseline works, including one that excludes the entity coherence model.
    Table~\ref{tab:crcn-performance} shows the results for the NY City~\cite{NIPS2015_17e62166} and Disneyland~\cite{NIPS2015_17e62166} datasets.
    Both datasets were introduced by Park et al.~\cite{NIPS2015_17e62166} and consist of image sequences with the corresponding blog posts.
    In these datasets, results were superior to all baselines. The exception, where it was a close second, was with the baseline without entity coherence and in only certain metrics.
    The work was further validated by user studies conducted via Amazon Mechanical Turk (AMT),
    showing that human evaluators prefer \gls{crcn}-generated sequences over several baselines.

    \begin{table}[htpb]
        \centering
        \begin{tabular}{l|ccccccccc}
            \hline
            Metric     & B-1                       & B-2                      & B-3                      & B-4  & C                        & M    & R@1                       & R@5                       & R@10                      \\
            \hline
            NY City    & 26.83                     & 5.37                     & \cellcolor{green!20}2.57 & 2.08 & 30.9                     & 7.69 & \cellcolor{green!20}11.67 & \cellcolor{green!20}31.19 & \cellcolor{green!20}43.57 \\
            Disneyland & \cellcolor{green!20}28.40 & \cellcolor{green!20}6.88 & \cellcolor{green!20}4.11 & 3.49 & \cellcolor{green!20}52.7 & 8.78 & \cellcolor{green!20}14.29 & \cellcolor{green!20}31.29 & \cellcolor{green!20}43.20 \\
            \hline
        \end{tabular}
        \caption[Performance metrics of \gls{crcn}]{Performance metrics of \gls{crcn} on New York City and Disneyland Datasets.
        Green values represent state-of-the-art results at the time. B stands for \gls{bleu}, C for \gls{cider}, M for \gls{meteor} and R for Recall~\protect\cite{NIPS2015_17e62166}.
        }
        \label{tab:crcn-performance}
    \end{table}

\begin{figure}[htpb]
        \centering
        \includegraphics[width=\textwidth]{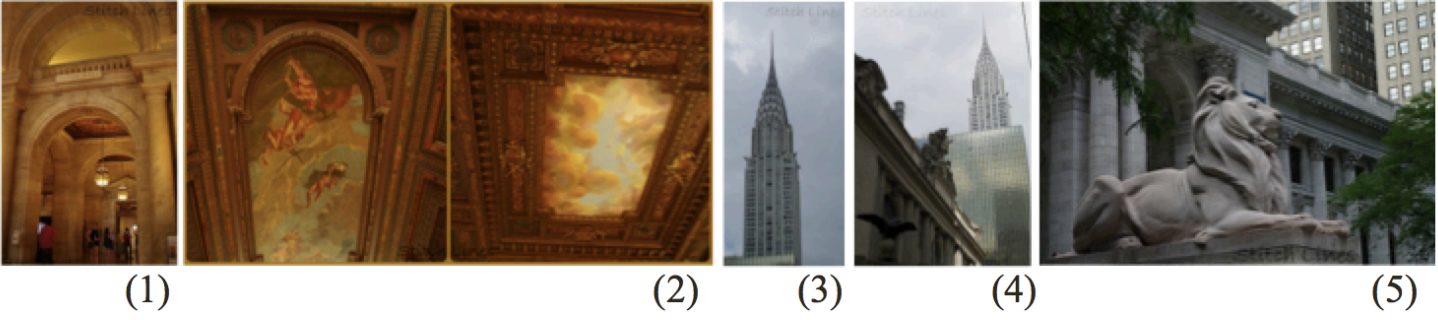}
        \caption{
            Sequence of images used by \gls{crcn} in a story.
            \textbf{(1)} One of the hallway arches inside of the library
            \textbf{(2)} As we walked through the library I noticed an exhibit called lunch hour nyc it captured my attention as
            I had also taken a tour of nyc food carts during my trip
            \textbf{(3)} Here is the top of the Chrysler building everyone's favorite skyscraper in new york.
            \textbf{(4)} After leaving the nypl we walked along 42nd st.
            \textbf{(5)} We walked down fifth avenue from rockefeller centre checking out the windows in saks the designer
            stores and eventually making our way to the impressive new york public library. Adapted from~\cite{NIPS2015_17e62166}.
        }
        \label{fig:crcn-example}
    \end{figure}

    Fig.~\ref{fig:crcn-example} shows an example of a sequence of images and the corresponding generated sentences.

    In addition to the work by Park et al.~\cite{NIPS2015_17e62166}, which was the first to use a bidirectional
    \gls{rnn} for visual storytelling, other works have also employed bidirectional \glspl{rnn} architectures.
    XE~\cite{wang-etal-2018-metrics} and AREL~\cite{wang-etal-2018-metrics} have proposed models for visual
    storytelling that use bidirectional \glspl{rnn} on the VIST dataset.

    \subsection{Transitional Adaptation of Pretrained Models for Visual Storytelling}
    \label{subsec:transitional-adaptation-of-pretrained-models-for-visual-storytelling}

    Yu et al.~\cite{Yu_2021_CVPR} proposed \gls{tapm}, aimed at refining the generation of textual
    descriptions for visual content, particularly in the context of \gls{vsg} tasks.
    It aims at bridging the gap between pretrained language models and visual encoders.
    In contrast with previous models for vision-to-language generation tasks,
    which typically pretrain a visual encoder and a language generator separately and then jointly fine-tune them for the target task,
    \gls{tapm} proposes a transitional adaptation task to harmonize the visual encoder and language model for downstream tasks like \gls{vsg},
    mitigating discord between visual specificity and language fluency arising from separate training on large corpora of visual and text data.
    \gls{tapm} introduces an approach that adapts multimodal modules through a simpler alignment task focusing solely on visual inputs, eliminating the need for text labels.

    \gls{tapm} components aim at enhancing the quality of textual descriptions for visual content in storytelling tasks. They are a visual encoder, a language generator, adaptation training, sequential coherence loss, training with adaptation loss, and a fine-tuning and inference process.
    The visual encoder, a pretrained model, extracts features from images or videos. In \gls{tapm}, it becomes integral during the adaptation phase, where it integrates with the language generator to fuse visual and textual information.
    The pretrained language generator model is responsible for converting visual information into textual descriptions.
    During the adaptation phase, it generates video and text embeddings, aligning textual representations with the corresponding visual features, based on a sequential coherence loss function.

    The loss function divides sequential coherence into three components: past, current, and future matching losses.
    The past matching loss uses a \gls{fc} layer \( f_p \) to project the text representation \( b_{s_i} \) of video \( i \), drawing it nearer to the visual representation \( v_{b_{i-1}} \) of the preceding video \( i - 1 \) and distancing
    it from those of non-sequential videos.
    The future matching loss projects \( b_{s_i} \) via a distinct \gls{fc} layer \( f_f \), aligns with the subsequent visual representation \( v_{b_{i+1}} \).
    The current matching loss then aligns the current visual representation \( v_{b_i} \) with \( b_{s_i} \) through another \gls{fc} layer \( f_c \). These components are unified by the \gls{fc} layer projections in their respective visual spaces, pulling the embeddings of correct matches closer and pushing incorrect matches further apart.
    Margin ranking losses are utilized to implement this concept, contrasting correct matches with incorrect ones.
    The final sequential coherence loss for a given video \( i \) is formulated as shown in Eq.~\ref{eq:sequential-coherence-loss}, where \( \cos \) is the cosine similarity, and \( j \) represents indices of incorrect matches.

    \begin{equation}
        \begin{aligned}
            L_i  = &\sum_{j \neq i-1} \max(0, 1 + \cos(v_{b_j}, f_p(b_{s_i})) - \cos(v_{b_{i-1}}, f_p(b_{s_i}))) \\
            + &\sum_{j \neq i} \max(0, 1 + \cos(v_{b_j}, f_c(b_{s_i})) - \cos(v_{b_i}, f_c(b_{s_i}))) \\
            + &\sum_{j \neq i+1} \max(0, 1 + \cos(v_{b_j}, f_f(b_{s_i})) - \cos(v_{b_{i+1}}, f_f(b_{s_i}))),
            \label{eq:sequential-coherence-loss}
        \end{aligned}
    \end{equation}

    \gls{tapm} uses a split-training strategy to optimize model performance.
    Initially, the visual encoder undergoes adaptation training with the adaptation loss, while the text encoder and language generator remain fixed.
    Subsequently, all components are jointly updated with the generation loss, allowing the model to optimize the adaptation task before addressing the more challenging generation objective.
    After the adaptation and split-training phases, \gls{tapm} undergoes fine-tuning.
    The model is then ready for the inference phase, generating captions for unseen visual inputs.

    Table.~\ref{tab:tapm-evaluation-results-vist} and~\ref{tab:tapm-evaluation-results-lsmdc} shows results for \gls{tapm} against selected baselines on the LSMDC 2019 and VIST datasets.
    Table.~\ref{tab:tapm-human-evaluation-lsmdc} show human evaluation results in which \gls{tapm} surpasses adversarial baselines on LSMDC 2019.
    On VIST it surpasses the XE~\cite{wang-etal-2018-metrics} and AREL~\cite{wang-etal-2018-metrics} baselines in relevance, expressiveness, and concreteness as shown in Table~\ref{tab:tapm-human-evaluation-vist}.
    These results highlight the strengths in word choice and contextual accuracy, showcasing its ability to capture causal relationships between images.
    However, the score is still far from human performance, indicating that there is still room for improvement.

    \begin{table}[htpb]
        \centering
        \begin{minipage}[t]{0.48\textwidth}
            \centering
            \begin{tabular}{l|ccc}
                \hline
                Models                                     & C                                 & M                                 & R                                 \\
                \hline
                Huang et al.~\cite{huang-etal-2016-visual} & -                                 & \cellcolor{red!20}31.4            & -                                 \\
                AREL~\cite{wang-etal-2018-metrics}         & \cellcolor{red!20}9.4             & 35.0                              & \cellcolor{red!20}29.5            \\
                StoryAnchor~\cite{modi-parde-2019-steep}   & 9.9                               & 35.5                              & 30.0                              \\
                HSRL~\cite{huang2018HierarchicallySR}      & 10.7                              & 35.2                              & 30.8                              \\
                INet~\cite{jung2020HideandTellLT}          & 10.0                              & 35.6                              & 29.7                              \\
                \gls{tapm}                                 & \cellcolor{green!20}\textbf{13.8} & \cellcolor{green!20}\textbf{37.2} & \cellcolor{green!20}\textbf{33.1} \\
                \hline
            \end{tabular}
            \caption{
                Quantitative results on the VIST test set. C stands for CIDEr, M for METEOR, and R for ROUGE-L.
                Green indicates the highest score, red the lowest.
            }
            \label{tab:tapm-evaluation-results-vist}
        \end{minipage}
        \hfill
        \begin{minipage}[t]{0.48\textwidth}
            \centering
            \begin{tabular}{l|cccc}
                \hline
                Models & \multicolumn{2}{c}{Public Test} & \multicolumn{2}{c}{Blind Test} \\
                & C                        & R                        & C                       & R                        \\
                \hline
                Baseline~\cite{park2018AdversarialIF} & \cellcolor{red!20}7.0    & 12.0                     & \cellcolor{red!20}6.9 & \cellcolor{red!20}11.9 \\
                XE~\cite{wang-etal-2018-metrics}      & 7.2                      & 11.5                     & -                       & -                        \\
                AREL~\cite{wang-etal-2018-metrics}    & 7.3                      & \cellcolor{red!20}11.4   & -                       & -                        \\
                \gls{tapm}                            & \cellcolor{green!20}10.0 & \cellcolor{green!20}12.3 & \cellcolor{green!20}8.8 & \cellcolor{green!20}12.4 \\
                \hline
            \end{tabular}
            \caption{
                Quantitative results on LSMDC 2019 public and blind test set. C stands for \gls{cider} and R for \gls{rouge}.
                Green indicates the highest score, red the lowest.
                Adapted from~\cite{Yu_2021_CVPR}.
            }
            \label{tab:tapm-evaluation-results-lsmdc}
        \end{minipage}\label{tab:tapm-evaluation-results}
    \end{table}

    \begin{table}[htpb]
        \centering
        \begin{tabular}{l|c}
            \hline
            Models                                         & Scores                    \\
            \hline
            Human                                          & \cellcolor{green!20}1.085 \\
            Official Baseline~\cite{park2018AdversarialIF} & \cellcolor{red!20}4.015   \\
            \gls{tapm}                                     & 3.670                     \\
            \hline
        \end{tabular}
        \captionsetup{width=\textwidth}
        \caption{
            Human evaluation results on the LSMDC 2019 blind test set according to a Likert scale from 5 (worst) to 1 (best), where lower is better.
            Green indicates the highest score, red the lowest.
            Adapted from~\cite{Yu_2021_CVPR}.
        }\label{tab:tapm-human-evaluation-lsmdc}
    \end{table}

    \begin{table}[htpb]
        \centering
        \begin{tabular}{l|ccc|ccc}
            \hline
            & \multicolumn{3}{c|}{\gls{tapm} vs XE} & \multicolumn{3}{c}{\gls{tapm} vs AREL} \\
            Choice (\%)    & \gls{tapm}               & XE                     & Tie  & \gls{tapm}               & AREL                   & Tie  \\
            \hline
            Relevance      & \cellcolor{green!20}59.9 & \cellcolor{red!20}34.1 & 6.0  & \cellcolor{green!20}61.3       & \cellcolor{red!20}32.8 & 5.9  \\
            Expressiveness & \cellcolor{green!20}57.3 & \cellcolor{red!20}32.3 & 10.4 & \cellcolor{green!20}57.3       & \cellcolor{red!20}34.0 & 8.7  \\
            Concreteness   & \cellcolor{green!20}59.1 & \cellcolor{red!20}30.3 & 10.7 & \cellcolor{green!20}59.6       & \cellcolor{red!20}30.4 & 10.0 \\
            \hline
        \end{tabular}

        \captionsetup{width=\textwidth}
        \caption[Human evaluation results of \gls{tapm} on VIST]{
            Human evaluation results on VIST. Higher is better.
            Green indicates the highest score, red the lowest.
            Adapted from~\cite{Yu_2021_CVPR}
        }
        \label{tab:tapm-human-evaluation-vist}
    \end{table}


    \subsection{Interactive and Creative Visual Storytelling}
    \label{subsec:interactive-and-creative-visual-storytelling}
    Building upon the foundational \gls{cnn} and \gls{rnn} approaches, interactive and creative visual storytelling represents
    an important evolution in the field, introducing intermediate approaches that bridge visual comprehension
    with narrative creativity.
    These methods combine elements of image captioning and \gls{vqa} while introducing narrative structure and creative
    interpretation, addressing limitations of purely technical approaches.

    Lukin et al.~\cite{lukin-etal-2018-pipeline} introduced a three-module pipeline for creative visual storytelling
    that systematically approaches the challenge of generating narratives from visual content.
    Their pipeline consists of object identification, single-image inferencing, and multi-image narration, that serve as a
    preliminary design for building a creative visual storyteller.
    The approach defines computational creative visual storytelling as one with the ability to alter the telling of a
    story along three aspects: to speak about different environments, to produce variations based on narrative goals,
    and to adapt the narrative to the audience.
    This modular approach demonstrates how the foundational capabilities of image captioning and \gls{vqa} can be
    systematically extended and combined to support more complex narrative generation tasks.
    The pipeline explicitly separates visual understanding from narrative construction,
    allowing for more systematic development and evaluation of each component while maintaining the overall
    coherence required for storytelling.

    Interactive visual storytelling has also explored user engagement and personalization in narrative generation.
    Halperin and Lukin~\cite{halperin2023EnvisioningNI} examined how creative visual storytelling can serve as an
    anthology for narrative intelligence, investigating the intersection of human creativity and automated
    story generation through an analysis of 100 visual stories from authors who participated in a
    systematic creative process of improvised story building based on image sequences.
    Their work on surreal visual storytelling~\cite{halperin2024ArtificialDS} further explores how visual narrative
    systems can handle ambiguous or dreamlike imagery, investigating AI ``hallucination'' by stress-testing a
    visual storytelling algorithm with different visual and textual inputs designed to probe dream logic inspired by
    cinematic surrealism.
    These approaches highlight the importance of modularity in visual storytelling systems, where different components
    can be optimized independently while contributing to the overall narrative coherence.
    The integration of creative elements with systematic visual analysis provides a foundation for more sophisticated \gls{vsg}
    systems that can balance factual visual description with imaginative narrative construction.

    \subsection{Knowledge-Enhanced Visual Storytelling}
    \label{subsec:knowledge-enhanced-visual-storytelling}

    A challenge in \gls{vsg} is the tendency of end-to-end approaches to produce monotonous
    stories with repetitive text and limited lexical diversity.
    This limitation arises because existing approaches are constrained by the vocabulary and knowledge available in
    single training datasets.
    To address this challenge, researchers have explored knowledge enhanced approaches that leverage external
    resources to enrich the story generation process.

    Hsu et al.~\cite{hsu2020knowledge} introduced KG-Story, a three-stage framework that allows \gls{vsg} systems to take
    advantage of external knowledge graphs to produce more diverse stories.
    The framework implements a distill-enrich-generate approach: first distilling a set of representative words
    from input prompts, then enriching the word set using external knowledge graphs, and finally generating stories
    based on the enriched word set.
    This framework allows the use of external resources not only for the enrichment phase, but also for the
    distillation and generation phases.
    The KG-Story framework operates through three distinct stages.
    In stage 1, an image-to-term model distills representative terms from each input image, creating conceptual
    representations that capture the essential elements of the visual content.
    In stage 2, external knowledge graphs are utilized to identify possible connections between the extracted
    term sets from different images, generating enriched term paths that capture relationships and associations
    not explicitly visible in the images.
    In stage 3, a Transformer architecture transforms these term paths into coherent stories,
    incorporating techniques such as length difference positional encoding and repetition penalties to
    improve narrative quality.
    Evaluation results demonstrate that stories generated by KG-Story are on average ranked better than previous state-of-the-art
    systems in human ranking evaluations.
    The approach successfully addresses the vocabulary limitations of traditional end-to-end methods
    while maintaining narrative coherence.

    Interactive approaches to knowledge enhanced storytelling have also been explored.
    Hsu et al.~\cite{hsu2019dixit} introduced Dixit, an interactive visual storytelling system that allows users to
    iteratively compose stories through term manipulation.
    The system extracts text terms describing objects and actions from photos, then allows users to add new terms or
    remove existing ones before generating stories based on these modified term sets.
    Behind the scenes, Dixit uses an \gls{lstm} model trained on image caption data to distill terms
    from images and utilizes a Transformer decoder to compose context coherent stories.
    This approach opens up possibilities for interpretable and controllable visual storytelling,
    allowing users to understand the story formation rationale and to intervene in the generation process.

    These knowledge enhanced approaches represent an important advancement in \gls{vsg}, demonstrating how
    external knowledge resources can be systematically integrated to overcome the limitations of purely
    data-driven methods.
    By explicitly modeling the relationship between visual content and broader conceptual knowledge,
    these systems can generate more diverse, engaging, and contextually rich narratives while maintaining
    controllability and interpretability.

    \subsection{Iterative and Planning-Based Approaches}
    \label{subsec:iterative-and-planning-based-approaches}
    Traditional \gls{vsg} models typically generate stories in a single forward pass.
    However, creative writers use their knowledge and worldview to put disjointed elements together to
    form a coherent storyline, and work and rework iteratively toward perfection.
    \gls{vsg} models, however, make poor use of external knowledge and iterative generation when attempting to
    create stories.
    This observation has motivated the development of iterative and planning-based approaches that more closely
    mirror human creative writing processes.
    Hsu et al.~\cite{hsu-etal-2021-plot} introduced PR-VIST (Plot and Rework Visual Storytelling), a
    framework that represents the input image sequence as a story graph in which it finds the best path to form a storyline.
    PR-VIST then takes this path and learns to generate the final story via a re-evaluating training process.
    The framework draws inspiration from the human creative writing process, which involves plotting
    (planning the overall narrative structure) and reworking (iteratively refining the story content).

    The PR-VIST framework operates through two main stages that correspond to the plotting and reworking phases of
    creative writing.
    In stage 1 (story plotting), the system constructs a story graph from the input image sequence using external
    knowledge graphs, including VIST and Visual Genome knowledge graphs.
    A storyline predictor model identifies the best path through this graph to form a coherent storyline,
    creating a structured narrative plan that connects the images through meaningful conceptual relationships.
    This plotting stage essentially creates a roadmap for the story that will be generated.
    In stage 2 (story reworking), the framework generates the actual story text based on the predicted storyline path.
    The approach uses a length controlled Transformer that is first pre-trained on the ROC Story dataset and then
    fine-tuned on the VIST dataset with a discriminator component to encourage higher quality story generation.
    The re-evaluating training process allows the model to iteratively improve the story generation by
    learning from feedback about narrative quality and coherence.

    Evaluation results demonstrate that this framework produces stories that are superior in terms of diversity,
    coherence, and humanness, per both automatic and human evaluations.
    An ablation study shows that both plotting and reworking contribute to the model's superiority.
    The explicit separation of planning and generation phases allows for better control over narrative structure
    while maintaining the flexibility needed for creative story generation.
    This planning based approach represents an important advancement toward more human-like story generation,
    demonstrating how computational systems can benefit from explicitly modeling the structured creative
    processes that human writers naturally employ.

    \subsection{Combining LLMs (GPT-2) and Character Features}
    \label{subsec:combining-llms-(gpt-2)-and-character-features}

\begin{figure}[htpb]
        \centering
        \includegraphics[width=\textwidth]{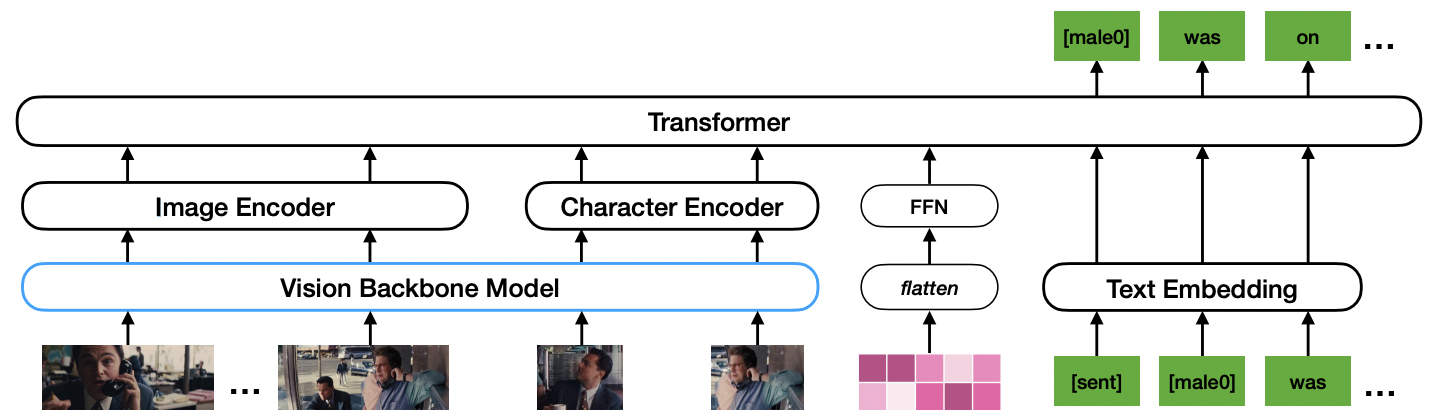}
        \caption{
            CharGrid architecture: the blue boxes are pre-trained components where the
            parameters are fixed. Adapted from~\cite{tacl_a_00553}.
        }
        \label{fig:char-grid-model}
    \end{figure}

    Hong et al.~\cite{tacl_a_00553}~proposed CharGrid (Character-Grid Transformer), a Transformer model
    that integrates diverse input features, from global image attributes to character-specific details.
    It takes a sequence of input tokens, including global image features obtained using Swin Transformer's~\cite{liu2021SwinTH},
    character features extracted by cropping character instances,
    a character grid for coherence assessment, and text features introduced incrementally during processing.
    The trainable image and character encoders process the image feature inputs.
    The character grid is flattened and fed to a \gls{fc} layer.
    The Transformer module then processes these inputs, generating an output as a probability distribution over potential tokens using a pre-trained \gls{gpt}-2 tokenizer.
    The model architecture is shown in Fig.~\ref{fig:char-grid-model}.

    During training, model parameters, excluding the vision backbone, are randomly initialized.
    The training objective involves maximizing the likelihood of image sequence-story pairs through backpropagation.
    Nucleus sampling~\cite{holtzman2019TheCC} is employed for validation, and the \gls{meteor} score is used for evaluation.

    CharGrid has two additional variants: ObjGrid (Object-Grid Transformer) and EntiGrid (Entity-Grid Transformer).
    ObjGrid replaces character features with object features. This model incorporates global features and object features, with an object grid representing coherence based on the similarity between global image features and object features.
    It is similar to CharGrid but includes both character and object features.
    EntiGrid incorporates global, character, and object features, with an entity grid representing coherence through the similarity between global image features and both character and object features.
    These variants explore the impact of different input feature combinations on model performance.
    CharGrid, with its emphasis on character coherence, serves as the primary model, while ObjGrid and EntiGrid
    address the contributions of object-related features.

    The results in Table~\ref{tab:char-grid-results} show that CharGrid outperforms \gls{tapm} with character features and \gls{gpt}-2 with character features on various metrics, emphasizing the effectiveness of character grid representations for coherence in \gls{vsg}.
    A crowd-sourcing experiment was conducted with 28 workers to obtain binary judgments on grammaticality, coherence, diversity, and visual groundedness
    of the generated stories.
    The results in Table~\ref{tab:chargrid-human-judgments} show that \gls{tapm} with character features excels in visual groundedness over plain \gls{tapm},
    while CharGrid surpasses \gls{tapm} with character features across all metrics.
    Two-sided binomial tests support that the character grid representation yields superior stories, affirming the reference-based metrics findings.

    \begin{table}[htpb]
        \centering
        \begin{tabular}{l|cccccccc}
            \hline
            Model                  & B-1                       & B-2                       & B-3                       & B-4                     & M                         & R-L                       & C                        \\
            \hline
            \gls{gpt}-2            & \cellcolor{red!20}38.65   & \cellcolor{red!20}20.28   & \cellcolor{red!20}9.78  & \cellcolor{red!20}4.68 & \cellcolor{red!20}31.64      & 24.24                     & 1.66                     \\
            \gls{gpt}-2 + obj      & 40.65                     & 21.35                     & 10.2                      & 4.87                    & 31.69                     & \cellcolor{red!20}24.05   & 1.85                     \\
            \gls{gpt}-2 + char     & 39.95                     & 21.04                     & 10.11                     & 4.92                    & 31.85                     & 24.19                     & 1.57                     \\
            \gls{gpt}-2 + obj,char & 40.41                     & 21.44                     & 10.56                     & 5.06                    & 32.03                     & 24.38                     & 1.87                     \\
            \gls{tapm}             & 39.85                     & 21.7                      & 10.72                     & 5.19                    & 32.38                     & \cellcolor{green!20}25.09 & 1.48                     \\
            \gls{tapm} + obj       & 40.86                     & 22.13                     & 10.83                     & 5.25                    & 32.34                     & 24.91                     & 1.82                     \\
            \gls{tapm} + char      & 40.03                     & 21.68                     & 10.66                     & 5.18                    & 32.42                     & 24.88                     & \cellcolor{red!20}1.4    \\
            \gls{tapm} + obj,char  & 40.87                     & 21.99                     & 10.72                     & 5.06                    & 32.48                     & 24.87                     & 1.59                     \\
            ObjGrid + obj          & 47.66                     & 25.26                     & 11.95                     & 5.42                    & 32.83                     & 24.42                     & 4.68                     \\
            EntityGrid + obj,char  & 45.83                     & 24.85                     & \cellcolor{green!20}12.11 & \cellcolor{green!20}5.7 & 32.68                     & 24.89                     & 3.53                     \\
            CharGrid + char        & \cellcolor{green!20}47.71 & \cellcolor{green!20}25.33 & 11.95                     & 5.42                    & \cellcolor{green!20}33.03 & 25.01                     & \cellcolor{green!20}4.83 \\
            \hline
        \end{tabular}
        \caption{Results using different input features on the test set of \gls{vwp}.
        Global features are always included. ``+ obj'' and ``+ char'' represent that
        object and character features are also included, respectively.
        \gls{bleu} (B), \gls{meteor} (M), \gls{rouge}-L (R-L), and \gls{cider} (C) values are averages of three runs
        with different random seeds.
        Green indicates the highest score, red the lowest.
        Adapted from~\cite{tacl_a_00553}.}
        \label{tab:char-grid-results}
    \end{table}

    \begin{table}[htbp]
        \centering
        \begin{tabular}{l|cccc}
            \hline
            Model                            & G       & C       & VG     & D        \\
            \hline
            \gls{tapm} + char vs. \gls{tapm} & +2.45   & +1.99   & +3.99* & +1.69    \\
            CharGrid vs. \gls{tapm} + char   & +6.49** & +8.41** & +6.25* & +11.06** \\
            \hline
        \end{tabular}
        \caption{Human binary judgments in percentage of generated stories between \gls{tapm} and \gls{tapm} with character features
            (\gls{tapm} + char), \gls{tapm} + char and CharGrid on the test set of \gls{vwp} across four criteria:
            Grammaticality (G), Coherence (C), Visual Groundedness (VG), and Diversity (D).
            The numbers are percentages. * means p-value $<$ 0.05. ** means p-value $<$ 0.01.
            Adapted from~\cite{tacl_a_00553}.
        }
        \label{tab:chargrid-human-judgments}
    \end{table}

    \subsection{Visual Story Generation using Graphs for Event Ratiocination}
    \label{subsec:visual-story-generation-using-graphs-for-event-ratiocination}

\begin{figure}[htpb]
        \centering
        \includegraphics[width=\textwidth]{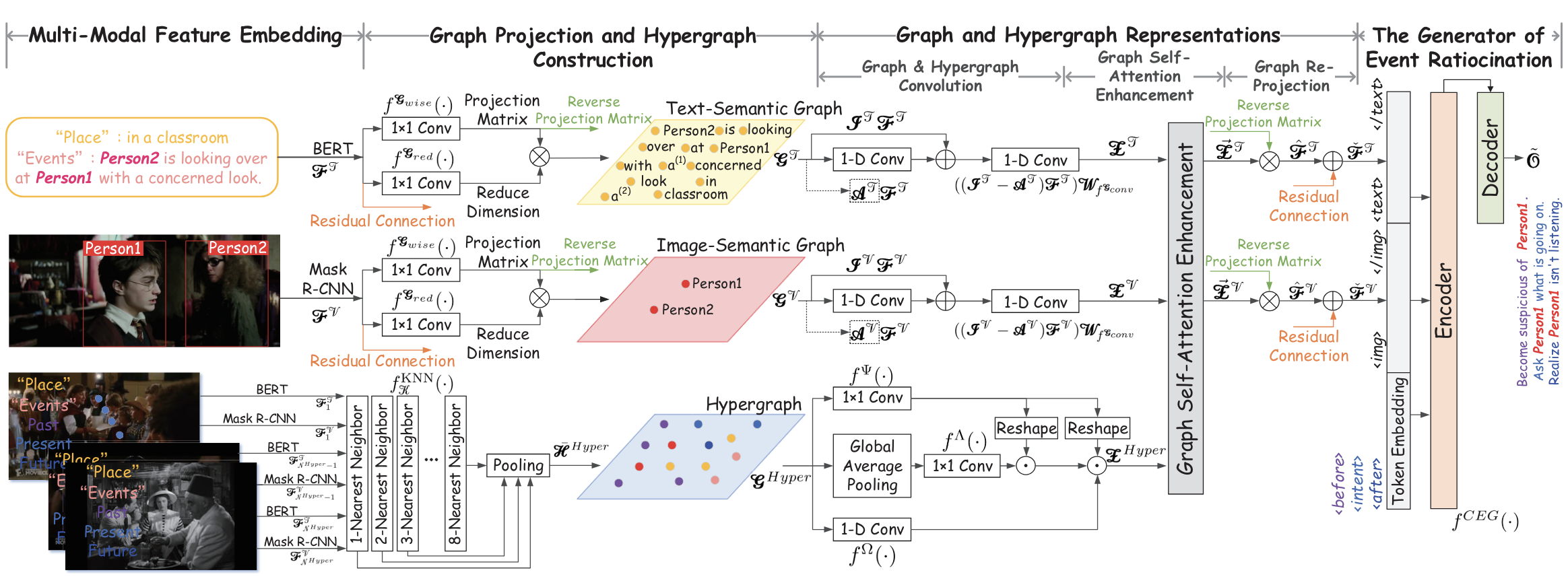}
        \caption{
            Architecture of HEGR by Zheng et al. Adapted from~\cite{pmlr-v139-zheng21b}.
        }
        \label{fig:two-heads-are-better-than-one-arquitecture}
    \end{figure}

    Zheng et al.~\cite{pmlr-v139-zheng21b} introduced Hypergraph-Enhanced Graph Reasoning (HEGR) for ``visual event ratiocination'', a task that involves generating interpretations and narratives for events that precede, coincide with, or follow a visual scene, as well as
    discerning the intents of depicted characters. The work addresses several downstream tasks, including \gls{vsg}.
    The work introduces a framework to improve the integration and interpretation of visual and textual data, using hypergraphs to address challenges in multimodal interactions and temporal dynamics. Fig.~\ref{fig:two-heads-are-better-than-one-arquitecture} shows this architecture.

    The method begins with feature extraction: visual features are extracted using Mask R-CNN~\cite{he2017MaskR} to
    detect objects and their spatial characteristics within images. Textual features are extracted from associated descriptions or captions using BERT~\cite{devlin2019BERTPO}, capturing semantic information.
    In the second phase,
    semantic graphs are constructed for each modality: nodes represent detected features or semantic tokens, while edges depict relationships within the same modality.
    To capture dependencies between modalities and across various temporal segments, hypergraphs are employed, linking semantic graphs based on contextual and temporal relevance, forming a high-dimensional interaction map.
    In the third phase,
    graph convolution is used to update node representations within each graph, allowing contextual information aggregation through learned adjacency matrices.
    Hyper-graph convolution extends this to include higher-order relationships across multiple graphs and temporal instances.
    In the fourth phase,
    a self-attention mechanism refines graph representations
    by dynamically adjusting the significance of nodes within and across modalities.
    This process focuses on the most relevant features for narrative generation.
    Finally, the event ratiocination generator uses the enriched feature representations to generate narratives using BART~\cite{lewis2019BARTDS}.

    The method was evaluated on the VIST dataset, achieving state-of-the-art performance in terms of \gls{bleu}-4, \gls{rouge}-L and \gls{meteor} scores achieving 16.4, 37.4, and 37.8, respectively.

    \subsection{Discussion}
    \label{subsec:discussion}

    The works presented in this section have made substantial contributions to the field of \gls{vsg}, establishing robust benchmarks for the assessment of emerging models and methodologies.

    Specifically, the first three techniques, presented in Sections~\ref{subsec:visual-storytelling-with-convolutional-and-recurrent-neural-networks},~\ref{subsec:visual-storytelling-with-convolutional-and-recurrent-networks}, and~\ref{subsec:transitional-adaptation-of-pretrained-models-for-visual-storytelling}
    employed a \gls{cnn} encoder for visual feature extraction and a \gls{rnn} decoder for language generation.
    The methods detailed in Sections~\ref{subsec:visual-storytelling-with-convolutional-and-recurrent-neural-networks} and~\ref{subsec:transitional-adaptation-of-pretrained-models-for-visual-storytelling},
    offered a comparative analyses on the VIST dataset, with METEOR scores of 31.4 and 37.2, respectively.
    Notably, the latter technique surpasses the former, which can be attributed to its transitional training approach.

    The study outlined in Section~\ref{subsec:combining-llms-(gpt-2)-and-character-features}, CharGrid,
    does not present findings on the VIST dataset.
    However, this method is benchmarked against the method detailed in Section~\ref{subsec:transitional-adaptation-of-pretrained-models-for-visual-storytelling}, \gls{tapm},
    on the \gls{vwp} dataset, where it demonstrates superior performance across all evaluated metrics.
    This comparison is shown in Table~\ref{tab:char-grid-results}, specifically within the rows labeled ``\gls{tapm} + obj,char'' and ``CharGrid''.
    This method differentiates itself from the previous by focusing on the extraction of object-level
    features from visual inputs rather than employing a \gls{cnn} to encode the entire visual input.
    Furthermore, it leverages the capabilities of an \gls{llm}, using its extensive world knowledge to enhance language modeling, illustrating a novel integration of visual and linguistic elements in story generation.

    Finally, the method discussed in Section~\ref{subsec:visual-story-generation-using-graphs-for-event-ratiocination}, HEGR,
    introduces a different approach that uses hypergraphs to model the contextual and temporal relationships between visual and textual data.
    This approach is explicitly able to model some story elements, namely the characters and the setting.

    Table~\ref{tab:visual-story-generation-methodologies} summarizes the performance of the methods described above,
    along with other methods from the literature that contributed to the field of \gls{vsg}.
    The listed models range from earlier methods using \glspl{cnn} and \glspl{rnn}
    to more recent advances that integrate object detection models like Faster R-CNN~\cite{ren2015FasterRT} and Mask-RCNN~\cite{he2017MaskR},
    and \glspl{llm} like GPT-2~\cite{radford2019LanguageMA} and BART~\cite{lewis2019BARTDS}.
    The first two methods by Park et al. (2015)~\cite{NIPS2015_17e62166} and Park et al. (2019)~\cite{park2018AdversarialIF}
    are not comparable to the other methods as they do not use the VIST dataset.
    The remaining methods are evaluated on the VIST dataset with the exception of CharGrid (2023)~\cite{tacl_a_00553}
    that is evaluated on the \gls{vwp} dataset.
    The authors of CharGrid trained and evaluated the performance of TAPM on the \gls{vwp}~\cite{tacl_a_00553} dataset,
    making the two methods directly comparable
    and it is likely that CharGrid would perform slightly better than TAPM on the VIST dataset.
    It is however not possible to know if CharGrid would perform better than HEGR on the VIST dataset
    without further experiments.

    \begin{sidewaystable}
        \centering
        \begin{tabular}{@{}lcccccccc@{}}
            \hline
            Model                                                       & Encoding                   & Decoding & Dataset     & B-4                      & R-L                       & M                         \\
            \hline
            Park et al. (2015)~\cite{NIPS2015_17e62166}                 & CNN, Bi-RNN                & RNN      & Blog Posts  & 3.49                     & -                         & 8.78                      \\
            \hline
            Park et al. (2019)~\cite{park2018AdversarialIF}             & Resnet-101/152, LSTM       & LSTM     & ActivityNet & 9.91                     & -                         & 16.48                     \\
            \hline
            Huang et al. (2015)~\cite{huang-etal-2016-visual}           & CNN, GRU                   & GRU      & VIST        & -                        & -                         & 31.4                      \\
            Xu et al. (2015)~\cite{xu2015ShowAA}                        & VGGNet                     & LSTM     & VIST        & -                        & \cellcolor{red!20}28.94   & 32.98                     \\
            Yu et al. (2017)~\cite{yu-etal-2017-hierarchically}         & ResNet-101, Bi-GRU         & GRU      & VIST        & -                        & 29.53                     & 34.12                     \\
            CST (2018)~\cite{gonzalezRico2018ContextualizeSA}           & CNN, LSTM                  & LSTM     & VIST        & 12.7                     & 29.2                      & 34.4                      \\
            AREL (2018)~\cite{wang-etal-2018-metrics}                   & ResNet-152, Bi-GRU         & GRU      & VIST        & 14.1                     & 29.5                      & 35.0                      \\
            HSRL (2019)~\cite{huang2018HierarchicallySR}                & ResNet-152                 & LSTM     & VIST        & \cellcolor{red!20}12.32  & 30.84                     & 35.23                     \\
            GLACNet (2019)~\cite{kim2018GLACNG}                         & ResNet-152, Bi-LSTM        & LSTM     & VIST        & -                        & -                         & \cellcolor{red!20}30.63   \\
            StoryAnchor (2020)~\cite{modi-parde-2019-steep}             & ResNet-152, Bi-GRU         & GRU      & VIST        & 14.0                     & 30.0                      & 35.5                      \\
            INET (2020)~\cite{jung2020HideandTellLT}                    & CNN, GRU                   & GRU      & VIST        & 14.7                     & 29.7                      & 35.5                      \\
            AOG-LSTM (2023)~\cite{liu2023AOG-LSTM}                      & RestNet-152                & LSTM     & VIST        & 12.9                     & 30.1                      & 36.0                      \\
            CoVS (2023)~\cite{gu2023coherent}                           & RestNet-152, GRU           & GRU      & VIST        & 15.2                     & 30.8                      & 36.5                      \\
            TAPM (2021)~\cite{Yu_2021_CVPR} \textsuperscript{1}         & Faster R-CNN, ResNet-152   & GPT-2 s   & VIST        & -                        & 33.1                      & 37.2                      \\
            HEGR (2021)~\cite{pmlr-v139-zheng21b} \textsuperscript{1,2} & Mask-RCNN, BERT            & BART     & VIST        & \cellcolor{green!20}16.4 & \cellcolor{green!20}37.4  & \cellcolor{green!20}37.8  \\
            \hline
            TAPM (2021)~\cite{tacl_a_00553}                             & Faster R-CNN, ResNet-152   & GPT-2    & \gls{vwp}   & \cellcolor{red!20}5.19   & \cellcolor{green!20}25.09 & \cellcolor{red!20}32.38   \\
            CharGrid (2023)~\cite{tacl_a_00553}\textsuperscript{2}      & Mask-RCNN, SwinTransformer & GPT-2    & \gls{vwp}   & \cellcolor{green!20}5.42 & \cellcolor{red!20}25.01   & \cellcolor{green!20}33.03 \\
            \hline
        \end{tabular}
        \caption{
            Analysis of Visual Story Generation works. The columns ``B-4'', ``R-L'', and ``M'' represent the \gls{bleu}-4, \gls{rouge}-L, and \gls{meteor} scores, respectively.
            All works, except Huang et al. (2015)~\cite{huang-etal-2016-visual} and CST (2018)~\cite{gonzalezRico2018ContextualizeSA}, use attention mechanisms. All works use images as input, except ActivityNet~\cite{caba2015activitynet} (it uses videos). The mark \textsuperscript{1} indicates works that used extra training data (direct comparison may not be fair). The mark \textsuperscript{2} indicates works that explicitly model some story elements, namely the characters by the detection of objects
            in the case of CharGrid (2023) and HEGR (2021) and the setting in the case of HEGR (2021).
        }
        \label{tab:visual-story-generation-methodologies}
    \end{sidewaystable}

    \section{Real-World Applications}\label{sec:real-world-applications}
    As \gls{vsg} technologies mature, they are increasingly being deployed in real-world scenarios.
    These applications demonstrate the practical potential of \gls{vsg} systems and highlight new challenges that
    arise when deploying these technologies in dynamic, context-rich environments.
    This section explores emerging applications that integrate visual storytelling with contextual information,
    location awareness, and real-time data processing.

    \subsection{Location-Aware Visual Storytelling}
    \label{subsec:location-aware-visual-storytelling}
    Suwono et al.~\cite{suwono2023location} introduced location-aware visual question generation with lightweight models,
    representing an important advancement in spatially-grounded narrative applications.
    Their work addresses the challenge of generating contextually relevant questions and narratives that are
    informed by geographical location, opening new possibilities for travel documentation,
    educational applications, and location-based social media platforms.
    The location-aware approach integrates geographical metadata with visual content analysis to create narratives
    that are visually grounded and spatially contextualized.
    This integration enables the generation of stories that can reference specific landmarks,
    cultural contexts, and geographical features that would be meaningful to users familiar with particular locations.
    The use of lightweight models addresses practical deployment constraints, making the technology suitable for mobile
    applications and resource-constrained environments.

    \subsection{Automotive and Real-Time Context Integration}
    \label{subsec:automotive-real-time-context}
    Belz et al.~\cite{belz2024story} explored story-driven approaches in automotive contexts,
    investigating how real-time contextual information can enhance automated storytelling systems.
    Their research represents a step toward practical deployment of \gls{vsg} in everyday scenarios,
    particularly in automotive systems where visual storytelling could enhance the driving experience and provide
    meaningful documentation of journeys.
    The automotive application domain presents challenges and opportunities for visual storytelling.
    In-car camera systems continuously capture visual data, creating an opportunity for real-time narrative
    generation that could help drivers and passengers understand and remember their journeys.
    The integration of real-time context information—such as speed, location, weather conditions,
    and traffic patterns enables the generation of richer, more informative narratives that extend beyond visual description.
    Their work demonstrates how dynamic contextual data can be incorporated into storytelling algorithms,
    showing measurable improvements in narrative relevance and user engagement when real-time information is provided.
    This approach could lead to applications in autonomous vehicles, where passengers might receive automatically
    generated travel narratives, or in fleet management systems where journey documentation becomes an automated process.

    \section{Conclusions}\label{sec:conclusions}
    This survey reviewed the field of \gls{vsg}, explaining its development and current methods.
    It started by explaining basic story elements, datasets and evaluation metrics used to judge the quality of the stories.
    The discussion pointed out the need for better metrics to assess complex tasks like \gls{vsg} accurately.
    The document also covered related areas like image and video captioning and \gls{vqa}, discussing how improvements in these fields have helped enhance \gls{vsg}.
    The discussion then moved on to the evolution of \gls{vsg} methods, starting with early approaches that used \glspl{cnn} and \glspl{rnn}
    to methods based on \glspl{llm} and hypergraphs.
    Additionally, we examined emerging real-world applications that demonstrate the practical potential of these technologies in domains such as automotive systems and location-aware storytelling.

    Despite the advances, challenges remain for \gls{vsg}.
    In the following subsections, we discuss some of the key challenges and opportunities in the field
    that could guide future research directions.
    We start with the ones that are more concrete and could influence the field in the short term and
    end with the ones that are more abstract and could potentially influence the field in the long term.

    \subsection{Exploring Large Language Models}
    \label{subsec:exploring-large-language-models}

    The field of \gls{vsg} using \glspl{llm} is under-explored, with only a few works delving into this domain.
    To the best of our knowledge the existing works predominantly rely on \gls{gpt}-2, which is an outdated \gls{llm} when compared to
    models such as Qwen3-235B-A22B~\cite{yang2025qwen3} which is the best performing open-source model on Chatbot Arena
    or Gemini 2.5 Pro~\cite{comanici2025gemini25} which is the best performing proprietary model.
    The scarcity of the integration of \glspl{llm} with visual elements presents an opportunity for further
    research, exploring the potential of these models in generating more coherent and contextually relevant stories.

    \subsection{Automatic Evaluation Metrics}
    \label{subsec:automatic-evaluation-metrics}

    The need for robust and efficient automatic evaluation metrics is emphasized in our survey.
    While existing metrics like \gls{meteor} and \gls{cider} have been widely adopted, they do not fully capture
    the nuances of story quality.
    The metrics mostly focus on the similarity between the generated text and the reference text.
    A story can be retold in various ways, employing different perspectives while preserving the original message.
    Consequently, a generated text might deviate significantly from the reference text in terms of specific wording and still
    deliver the same story in a coherent way.
    Thus, while these metrics provide valuable insights into certain dimensions of text generation, they did not fully
    encompass the multifaceted nature of storytelling.
    As \glspl{llm} continue to improve, they show potential in replacing human annotators in many tasks~\cite{gilardi2023ChatGPTOC}
    and so metrics to evaluate stories could be developed based on \glspl{llm}.
    We advance that these metrics could be based on the existing systematic scoring systems~\cite{zheng2023judging}
    that were developed to evaluate the performance of \gls{llm}-based chat bots.

    \subsection{Decomposition of Story Generation}
    \label{subsec:decomposition-of-story-generation}

    Addressing the complexity of stories, we underscore the importance of decomposing the story generation process into modular components.
    Each module can focus on specific attributes such as author goals, character interactions, consistency, etc.
    This decomposition has already been explored with modeling characters in CharGrid~\cite{Yu_2021_CVPR} and the
    modeling of characters and setting in HEGR~\cite{pmlr-v139-zheng21b}.
    We believe that explicit modeling of other story elements and abstract concepts such as emotions and culture,
    possibly using hypergraphs or similar structures could further enhance the quality of generated stories.

    \subsection{Hybrid Systems and Knowledge Integration}
    \label{subsec:hybrid-systems-and-knowledge-integration}

    Building on the decomposition concept, hybrid systems, combining planning-based techniques and machine learning approaches, present an intriguing direction.
    Hyper-graphs, as seen in HEGR~\cite{pmlr-v139-zheng21b}, could possibly be a starting point as a deep learning approach
    to model the planning process replacing the traditional handcrafted planning systems.
    The integration of multiple knowledge sources can enrich the diversity and quality of generated stories, paving the way for more robust and
    context-aware narrative creation.

    \subsection{Real-World Deployment and Context Integration}
    \label{subsec:real-world-deployment-context-integration}
    The transition from academic research to practical applications represents a crucial next step for \gls{vsg} technologies.
    Recent work in location aware storytelling and automotive applications demonstrates the potential for deploying these systems in real-world scenarios.
    Future research should focus on developing robust systems that can integrate multiple contextual information sources including location, temporal data,
    and real-time sensor information while maintaining computational efficiency for deployment in resource constrained environments such as
    mobile devices and embedded systems.
    The success of these real-world applications will ultimately determine the broader impact of \gls{vsg} research on society.

    \acknowledgments{%
        Daniel A. P. Oliveira is supported by a scholarship granted by \gls{fct},
        with reference 2021.06750.BD. Additionally, this work was supported by Portuguese national funds through \gls{fct},
        with reference UIDB/50021/2020.
    }

\isPreprints{}{
    \begin{adjustwidth}{-\extralength}{0cm}
} 
        \reftitle{References}

        \bibliography{bibliography}

        \PublishersNote{}
\isPreprints{}{
    \end{adjustwidth}
} 

\end{document}